\PassOptionsToPackage{table,xcdraw}{xcolor}
\documentclass[11pt]{article}
\usepackage[preprint]{acl}
\usepackage{subcaption}
\usepackage{times}
\usepackage{latexsym}
\usepackage[T1]{fontenc}
\usepackage[utf8]{inputenc}
\usepackage{microtype}
\usepackage{inconsolata}
\usepackage{graphicx}
\usepackage{booktabs}
\usepackage{amsmath,amsfonts,amssymb}
\usepackage{multirow}

\title{ChiEngMixBench: Evaluating Large Language Models on Expert-Style Chinese-English Terminology Mixing}

\author{Qingyan Yang \\
  School of Future Technology\\
  Southeast University \\
  Nanjing, China \\\And
  Tongxi Wang\thanks{Corresponding author} \\
  School of Future Technology\\
  Southeast University \\
  Nanjing, China \\
  \texttt{tongxi\_wang@seu.edu.cn} \\\And
  Yunsheng Luo \\
  School of Future Technology\\
  Southeast University \\
  Nanjing, China}

\begin{document}
\maketitle

\begin{abstract}
Large language models increasingly mediate multilingual professional communication, where useful generation requires adapting to community conventions about which expressions are retained, translated, or mixed. Existing benchmarks rarely isolate such community-conditioned choices. We introduce \textit{ChiEngMixBench}, a controlled benchmark for Chinese AI/CS discourse, where Chinese frames routinely incorporate established English technical terms. Built from public technical discussions, it contains 1,706 source-derived candidate pairs covering 1,344 non-empty normalized terms, including a 1,167-pair strict subset that fixes the Chinese prefix and syntactic position while varying only the terminology form. The benchmark combines paired likelihood comparisons with a transparent reference-profile diagnostic for open-ended responses. Across nine open-weight models, Chinese equivalents receive higher likelihood on most pairs, revealing a gap between source-attested usage and model preference. Specialized terms show a small directional lift that is not robust after frequency and length controls and multiple-comparison correction. Human evaluation and baseline analyses show that reference-profile conformity is informative under the intended mixed-style rubric but does not reliably predict holistic response preference. \textit{ChiEngMixBench} provides a reusable testbed for community-specific multilingual conventions with explicit diagnostic boundaries.

\end{abstract}

\section{Introduction}
\label{sec:intro}

Large language models are increasingly used not only to translate or retrieve information, but also to participate in multilingual professional communication. In such settings, a useful response must be factual and fluent, but it must also choose expressions in ways that are recognizable to its intended community. Professional groups develop local conventions around abbreviations, borrowed terminology, untranslated names, and language mixing. These conventions affect whether an answer reads as concise, familiar, and usable even when competing forms are semantically equivalent. Evaluating adherence to such conventions is therefore a distinct aspect of multilingual model behavior, rather than a proxy for translation accuracy or factual knowledge.

Chinese AI and computer-science discourse offers a particularly informative setting in which to study this problem. Practitioners frequently embed established English terms such as ``transformer'', ``fine-tuning'', and ``logits'' in otherwise Chinese sentences. This practice is neither arbitrary substitution nor necessarily a failure to translate. It is a conventional form of insertional code-mixing in which Chinese supplies the morphosyntactic frame and English supplies selected technical expressions, broadly consistent with the Matrix Language Frame account \cite{myers1993duelling}. A response that translates every term into Chinese may be grammatical and understandable yet depart from familiar community practice; a response that inserts English indiscriminately may be equally inappropriate. This setting therefore provides a tractable instance of a broader evaluation question: can a model adapt its lexical choices to a community whose preferred forms are observable in naturally occurring text? We do not assume that these conventions generalize to all domains. Instead, their recurrence and the abundance of public technical discussion make controlled evaluation possible.

\begin{figure*}[t]
    \centering
    \includegraphics[width=0.90\textwidth]{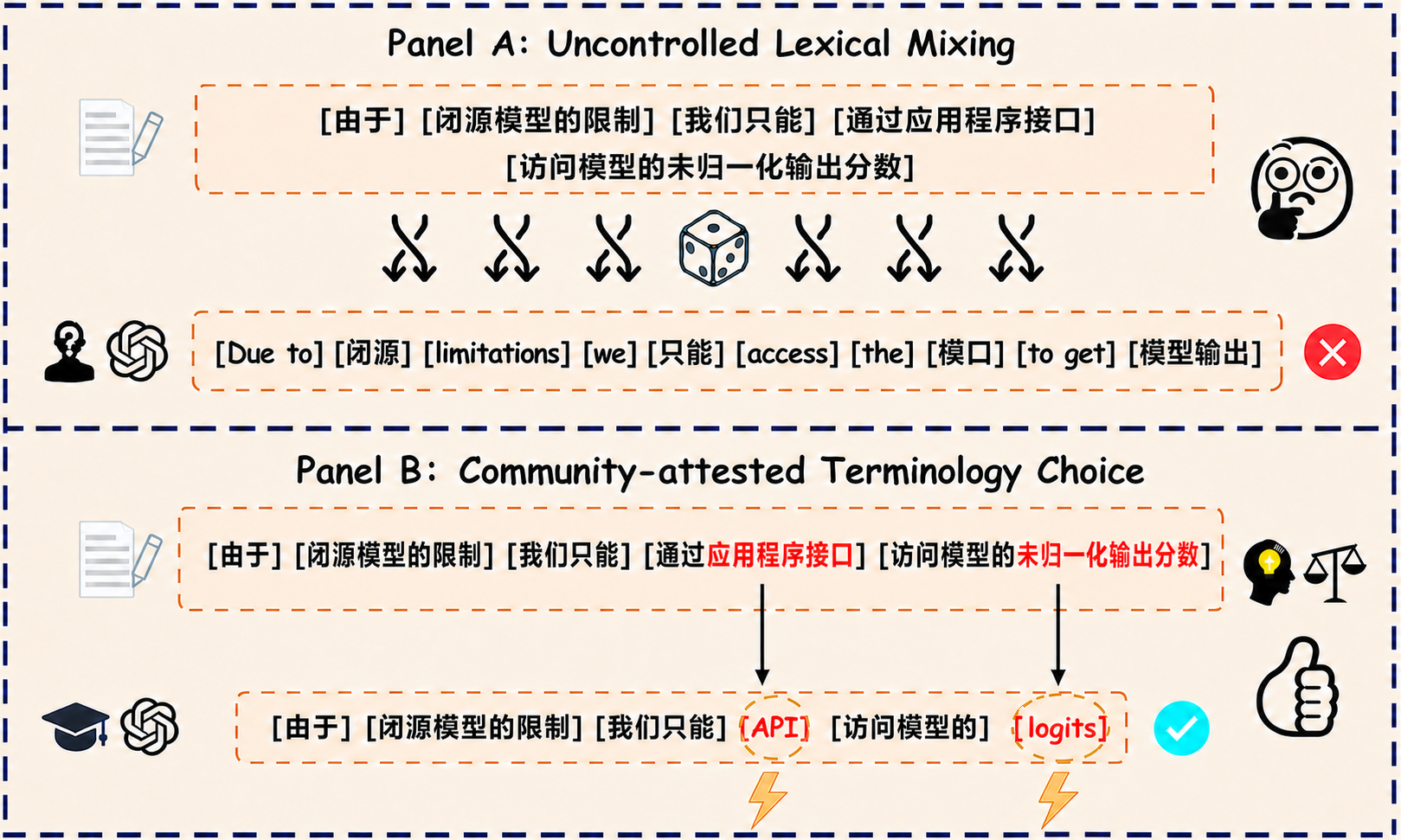}
    \caption{\textbf{Task intuition of ChiEngMixBench.} Panel A illustrates uncontrolled lexical mixing, which may interleave languages without preserving readable community practice. Panel B illustrates community-attested terminology choice: established technical forms such as \textit{API} and \textit{logits} are retained within a Chinese sentential frame. The benchmark does not reward English insertion per se; it evaluates whether terminology choices match source-attested usage under controlled contexts.}
    \label{fig:intro_concept}
\end{figure*}

Existing multilingual and code-mixing benchmarks provide valuable coverage of translation, language identification, cross-lingual understanding, and mixed-language generation. They are less suited, however, to isolating a model's choice between two terminology forms under an otherwise identical context. Naturally occurring text offers ecological relevance but introduces variation in topic, syntax, and discourse history; synthetic word replacement offers control but may not reflect established community usage. Evaluating community-conditioned language choice therefore requires both source-attested evidence and controlled comparisons, as well as a way to distinguish local lexical preference from the behavior of a complete generated response. Figure~\ref{fig:intro_concept} illustrates the distinction between uncontrolled lexical mixing and community-attested terminology choice.

To make this form of community-conditioned choice measurable, we introduce \textit{ChiEngMixBench}, a benchmark derived from public Chinese AI/CS technical discussions. The resource contains 1,706 source-derived candidate pairs covering 1,344 non-empty normalized terms, of which 1,167 form the strict single-replacement primary set. Within each strict pair, the English term and a verified Chinese equivalent occur after the same Chinese prefix and occupy the same syntactic position. This design retains evidence of actual community usage while controlling much of the topical and structural variation that would otherwise confound a comparison between the two forms. Because source-user demographics are unavailable, we describe the target distribution as \textit{expert-style technical discourse}, rather than treating all source text as an inherently representative expert corpus.

ChiEngMixBench supports two complementary observations rather than collapsing model behavior into a single ranking. The first, \textit{paired lexical preference}, compares length-normalized log likelihoods for the English term and its Chinese equivalent within each controlled pair. It measures an observable probability preference without inferring an unobserved cognitive mechanism. The second, \textit{generation conformity}, analyzes open-ended responses relative to a corpus-derived mixed-style reference profile. Its Expert Deviation Penalty (EDP) decomposes observable deviations in mixing ratio, switching structure, semantic style, and terminology form. EDP is an auditable, task-specific diagnostic rather than a universal measure of response quality or linguistic naturalness. We further test what these observations do and do not establish through blind human evaluation, alternative aggregation rules, an English-ratio baseline, negative controls, statistical uncertainty, and frequency/length controls.

The resulting evidence reveals a consistent mismatch, but also places clear limits on its interpretation. Across the nine evaluated open-weight models, Chinese equivalents receive higher likelihood on most matched pairs even though the English terminology is attested in the source communities. Specialized terms exhibit a small, directionally consistent lift in relative likelihood, but this pattern is not robust after multiple-comparison correction and frequency/length controls. Open-ended generation behavior varies substantially across the 12 evaluated systems. EDP is strongly associated with human ratings collected under the same mixed-style rubric and provides better calibration and decomposable error information than an English-ratio-only baseline. However, a separate blind pairwise study shows that it does not reliably predict holistic human preference for complete responses. We therefore interpret it only as reference-profile conformity and report model comparisons descriptively; the available Base/Instruct comparison does not support a general claim about instruction tuning or RLHF.

The resulting benchmark turns a common but poorly measured communicative practice into a controlled evaluation setting. It anchors candidate terminology choices in public community usage, separates local lexical preference from full-response behavior, and accompanies its diagnostics with explicit reliability and scope checks. This combination makes ChiEngMixBench reusable both for comparing models and for examining where otherwise fluent output departs from the conventions of a target community. More broadly, it provides a concrete methodology for studying community-conditioned multilingual generation without treating the preferences of one community as universal linguistic norms.

\paragraph{Reproducibility Statement}
The anonymous supplementary material contains the benchmark data, evaluation code, configurations, and command-line instructions. We plan to release the maintained resource and software publicly upon publication.

\section{Related Work}
\label{sec:related_work}

\subsection{Code-Mixed Data and Evaluation}
Research on code-mixing has progressed from conversational speech corpora such as SEAME \citep{lyu2010seame} to social-media and benchmark datasets covering multiple language pairs and tasks \citep{aguilar2020lincecentralizedbenchmarklinguistic,srivastava-singh-2020-phinc,khanuja-etal-2020-gluecos,winata-etal-2023-decades}. These resources have supported work on language identification, understanding, translation, and generation. However, many benchmarks emphasize informal conversation or broad multilingual competence, whereas terminology choices in technical communities are conditioned by domain conventions and by the surrounding matrix language. Synthetic construction through random or grammar-constrained substitutions is useful for scale and control \citep{pratapa-etal-2018-language,rizvi-etal-2021-gcm}, but may not preserve which English forms practitioners conventionally retain in a particular technical context.

Recent benchmarks broaden code-mixed evaluation across languages and tasks \citep{yang2025codemixbench,sheokand2025codemixbench,shirke2025comparative}, and domain-specific studies have examined settings such as mixed clinical text \citep{li2023unlocking}. ChiEngMixBench is complementary: it focuses on Chinese-matrix AI/CS discussion and isolates English-versus-Chinese terminology choices under matched local contexts. We therefore treat the benchmark as a targeted resource rather than as a replacement for general code-mixing evaluation.

\subsection{Multilingual Behavior of Language Models}
Prior work evaluates whether multilingual models understand mixed inputs, follow output-language constraints, or generate code-mixed text when explicitly requested \citep{huzaifah2024,marchisio-etal-2024-understanding,lee-etal-2025-controlling,yong2023prompting}. These settings primarily measure task success under an instruction. Our paired evaluation instead asks which of two matched sentence realizations, differing in the target terminology form, receives higher length-normalized likelihood. A separate open-ended evaluation then records how generated answers conform to a task-specific reference profile. This separation avoids using generation behavior as a direct proxy for matched-pair likelihood preference, while also avoiding the assumption that paired preference alone determines complete-response quality.

Output-language alignment research further shows that models may not consistently match the language expected by a user or context \citep{kim2026ola}. ChiEngMixBench studies a narrower version of this problem: matching terminology conventions within one technical community. We do not interpret the resulting behavior as a general measure of model-internal mechanisms or bilingual ability.

\subsection{Community-Conditioned Terminology Choice}
Sociolinguistic accounts such as the Matrix Language Frame describe structurally asymmetric mixing, in which one language supplies the grammatical frame and another supplies embedded material \citep{myers1993duelling}. Communities of practice also develop shared lexical repertoires \citep{lave1991situated}; in Chinese AI/CS discussion, compact English names and acronyms can become conventional within otherwise Chinese sentences. These theories motivate the task design, but they do not by themselves establish a causal mechanism for model behavior. Accordingly, our analyses use descriptive terms such as \emph{paired terminology-choice preference} and \emph{generation conformity}, and reserve theoretical explanations for discussion rather than treating them as experimentally verified mechanisms.

\section{Dataset Construction}
\label{sec:dataset}

\begin{figure*}[t]
    \centering
    \includegraphics[width=0.98\textwidth]{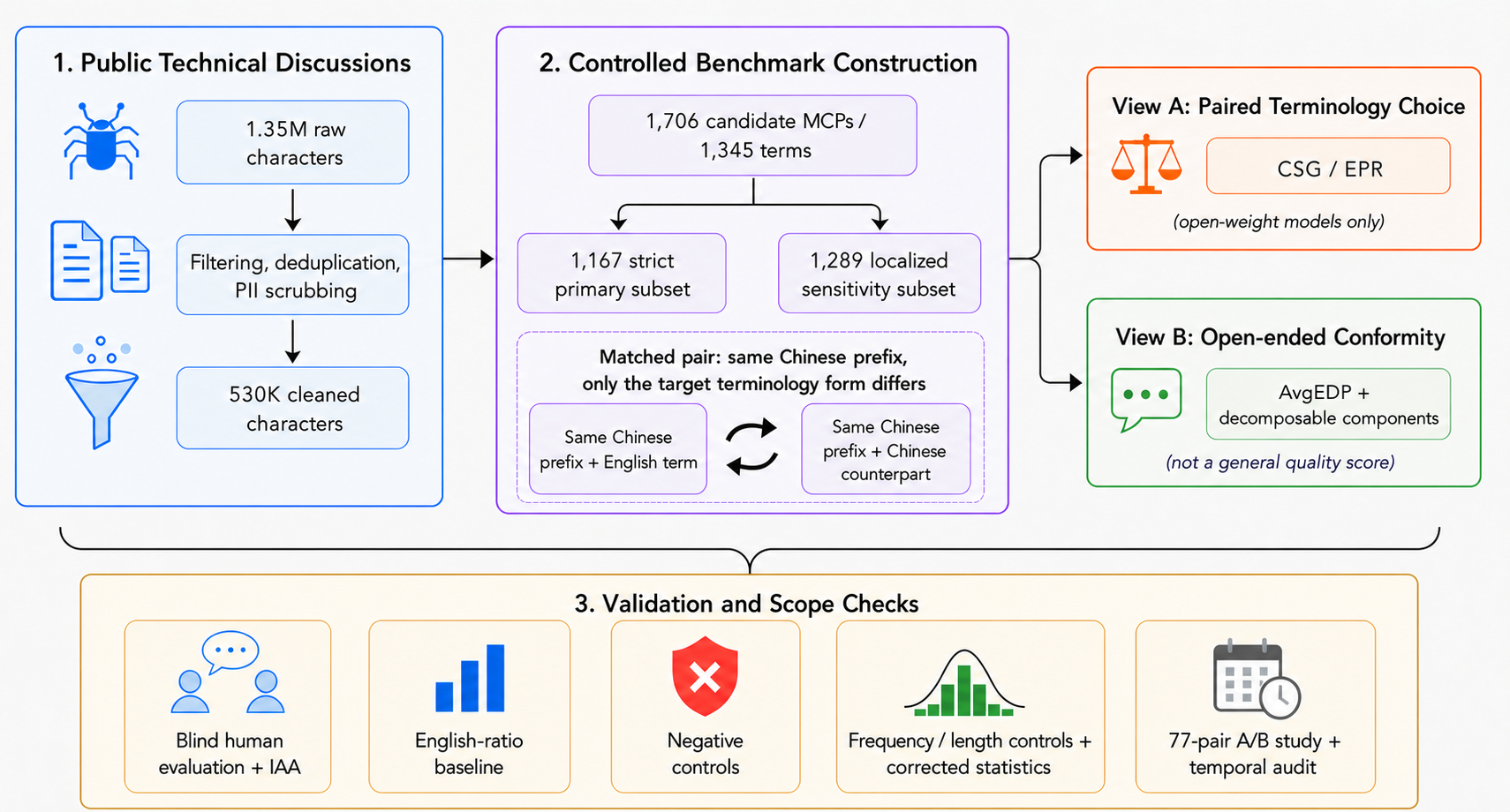}
    \caption{\textbf{ChiEngMixBench construction and audit pipeline.} Public Chinese technical discussions are filtered into a 530K-character working corpus, from which 1,706 candidate MCPs are constructed, with 1,344 non-empty normalized target terms. A deterministic pair-fidelity audit yields a 1,167-pair strict primary subset and a 1,289-pair localized-edit sensitivity subset. The benchmark then separates paired terminology choice for open-weight models from open-ended generation conformity and reports human-evaluation, baseline, negative-control, statistical-control, and temporal-audit checks that bound the supported interpretations.}
    \label{fig:pipeline}
\end{figure*}

ChiEngMixBench studies terminology mixing in Chinese-matrix AI/CS discourse. The released resource contains 1,706 source-derived candidate pairs and 1,344 non-empty normalized target-term types, together with pair-quality labels and a reference profile for open-ended generation diagnostics. To avoid treating every historical pair as an equally controlled contrast, the primary paired analysis uses a deterministic strict subset of 1,167 single-replacement pairs covering 934 normalized target terms. A broader 1,289-pair localized-edit subset is retained for sensitivity analysis. Figure~\ref{fig:pipeline} summarizes benchmark construction, the two evaluation views, and the validation checks used to delimit their interpretation.

\subsection{Source Scope and Processing}
\label{sec:data_acquisition}
We collected publicly accessible technical discussions from CSDN and Zhihu between May 2021 and November 2025. The collection focuses on AI and computer-science topics and contains approximately 1.35M characters before filtering. We applied exact deduplication, length and topic filters, terminology-density checks, and manual inspection of candidate passages. The resulting working corpus contains approximately 530K characters.

We refer to these data as \emph{expert-style technical discourse}: they exhibit terminology practices found in Chinese technical communities, but their public-web origin does not guarantee that every author is a domain expert. Author demographics were not available and are not inferred. The public release therefore focuses on processed, de-identified items and derived annotations rather than presenting the corpus as a demographically representative sample of all Chinese AI/CS practitioners.

\subsection{Candidate Pair Construction}
\label{sec:mcp_construction}
Each candidate pair contrasts an English technical term with a Chinese counterpart in a Chinese-matrix sentence. Candidate terms and contexts were extracted from the working corpus, and a bilingual terminology table supplied candidate counterparts. The intended construction holds topic and local context approximately fixed while changing the terminology realization. Because the historical collection process sometimes retained the target English abbreviation or term on the Chinese side, changed punctuation around the target span, or rewrote words outside the target terminology, we do not assume that all 1,706 candidates are minimal contrasts.

\subsection{Pair-Fidelity Audit and Analysis Splits}
\label{sec:pair_audit}
We apply a deterministic character-level audit before analysis. The audit checks whether the recorded English target occurs in the English-side sentence, is absent from the Chinese-side sentence, and is covered by the changed span between the two strings. A \emph{strict single-replacement} pair has one changed block and at most two off-target changed characters. A \emph{localized-edit} pair may contain up to three changed blocks and at most eight off-target changed characters. These labels are reproducible quality-control rules, not claims of human-verified linguistic perfection.

Across the 1,706 candidates, the audit assigns 1,168 to strict single replacement, 122 to localized replacement with minor surface edits, 294 to target retained on the Chinese side, 112 to non-local or multi-edit pairs, three to missing target metadata, and seven to identical pairs. Two empty target fields are recoverable directly from the English-side sentence and are restored without changing sentence text; one duplicate strict pair is excluded. The resulting primary analysis set contains 1,167 strict pairs, and the localized-edit sensitivity set contains 1,289 pairs. The complete 1,706-item collection remains available with quality labels so that future work can revise or study the excluded cases without silently mixing them into the primary estimates.

For the primary strict subset, 1,079 items are labeled \emph{specialized} and 88 \emph{general}. Specialized items include domain-specific acronyms, model or library names, methods, and technical expressions whose compact English form is conventional in AI/CS discussion. General items are cross-domain words with commonly used Chinese counterparts. This descriptive grouping is highly imbalanced (92.5\% versus 7.5\%) and is not a universal linguistic taxonomy. We therefore report uncertainty estimates, multiplicity corrections, and frequency/length controls rather than treating the subgroup contrast as a definitive effect (Section~\ref{sec:paired_results}).

\subsection{Reference Profile for Generated Text}
\label{sec:reference_profile}
The working corpus also provides task-specific reference statistics for open-ended answers: an English-token-ratio interval, a switching-frequency limit, a semantic embedding centroid, and a static set of terminology-form checks. These quantities define the reference profile used by the Expert Deviation Penalty (EDP; Section~\ref{sec:framework}). They are transparent, human-specified properties of this benchmark rather than universal boundaries for natural bilingual writing.

\subsection{Provenance and Release Boundaries}
The source platforms may contain reposted, search-optimized, or model-generated content, and text published before a model release may still overlap with unknown pretraining data. Within the strict primary subset, the paired design controls much of the topic and local-context variation, but it cannot prove the absence of differential exposure to the original mixed form. We document filtering, pair-quality labels, and release boundaries in the data statement, and report a small external temporal audit in Section~\ref{sec:scope_checks}. Because stable upstream document and author identifiers were not retained for every item, the current release does not support author-disjoint or document-disjoint splitting; this is an explicit limitation and a target for future data collection.

\section{Evaluation Protocol}
\label{sec:framework}

ChiEngMixBench uses two complementary views. The first measures relative sequence preference between the English-term and Chinese-counterpart realizations of a matched pair. The second summarizes selected properties of an open-ended answer relative to the benchmark's reference profile. The two views answer different questions and are not combined into a single model ranking. The paired lexical view is the benchmark's primary controlled evaluation; the open-ended view is an auxiliary diagnostic for inspecting generation failures, not a co-equal ranking axis.

\subsection{Paired Terminology-Choice Preference}
For MCP $j$, let $s_v^{(j)}=(x_1,\ldots,x_{L_v})$ denote the complete sentence realization for terminology form $v\in\{\mathrm{en},\mathrm{zh}\}$. We compute its length-normalized autoregressive log-likelihood
\begin{equation}
\label{eq:term_score}
\ell\!\left(s_v^{(j)}\right)=\frac{1}{L_v^{(j)}}\sum_{i=1}^{L_v^{(j)}}
\log P_{\theta}\!\left(x_i\mid x_{<i}\right),
\end{equation}
where tokenization and sequence length are model-specific. The Contextual Selection Gap (CSG) is
\begin{equation}
\label{eq:csg}
\mathrm{CSG}^{(j)}=\ell\!\left(s_{\mathrm{en}}^{(j)}\right)
-\ell\!\left(s_{\mathrm{zh}}^{(j)}\right).
\end{equation}
In the strict primary set, the two realizations differ in one localized replacement block whose recorded target is the intended contrast. A positive CSG indicates that the complete English-term realization receives higher average log-likelihood, whereas a negative value favors the Chinese-counterpart realization. This is a matched-sentence score rather than a direct readout of target-token probability, and length normalization does not remove all tokenization effects. We also report the English Preference Rate
\begin{equation}
\label{eq:epr}
\mathrm{EPR}=\frac{100}{N}\sum_{j=1}^{N}\mathbb{I}[\mathrm{CSG}^{(j)}>0].
\end{equation}
CSG is meaningful only within the paired construction and should not be read as an absolute probability of code-mixing. Closed APIs that do not expose stable sequence-level log-likelihoods are excluded from this view.

\subsection{Generation Conformity Diagnostic}
\label{sec:edp}
For open-ended answers, EDP records deviations from the task-specific reference profile. For item $i$,
\begin{align}
\label{eq:edp}
\mathrm{EDP}_{i}&=\max\left(0,5-\sum_{k}P_{ik}\right), \\
\overline{\mathrm{EDP}}&=\frac{1}{M}\sum_{i=1}^{M}\mathrm{EDP}_{i}.
\end{align}
The clipping is applied \emph{before} averaging. Consequently, means of individual penalty columns cannot be subtracted from 5 to reconstruct $\overline{\mathrm{EDP}}$.

The released scorer has four auditable components: English-token ratio, switching frequency, semantic distance from the released multilingual MiniLM reference centroid \citep{wang2020minilm}, and triggered terminology-form rules. Thus, the sum in Eq.~\ref{eq:edp} is $P_i^{\mathrm{ratio}}+P_i^{\mathrm{switch}}+P_i^{\mathrm{semantic}}+P_i^{\mathrm{term}}$. Appendix~\ref{app:edp_audit} reports the exact piecewise rules, thresholds, coefficients, and implementation audit. These are benchmark-specific design choices rather than universal criteria for naturalness. We release the centroid, terminology rules, machine-readable configuration, component outputs, and scoring code so that every deduction can be inspected.

\subsection{What the Two Views Establish}
The paired view compares strict matched sentence realizations whose controlled contrast is the terminology form. EDP instead describes how an answer compares with one mixed-style reference profile along decomposable dimensions. A model can favor Chinese-counterpart realizations in many pairs yet still produce a fluent answer, and a high EDP score does not imply that humans will prefer the complete response. We test these boundaries with same-rubric Likert ratings, a ratio-only baseline, a separate blind A/B preference study, and negative controls in Section~\ref{sec:validation_scope}.

\section{Experiments and Analysis}
\label{sec:experiments}

\subsection{Experimental Setup}
View A covers nine open-weight checkpoints on the 1,167-pair strict primary set; the 1,289-pair localized-edit set provides a sensitivity analysis. View B covers nine open-weight checkpoints and three closed API systems. The checkpoint sets are not fully identical across views: the paired analysis uses Meta-Llama-3-8B and Mistral-7B-v0.3, whereas the available generation logs use Meta-Llama-3.1-8B-Instruct and Mistral-7B-Instruct-v0.3. Closed systems are omitted from paired terminology-choice preference because their APIs do not provide stable sequence-level log-likelihoods under a comparable setup. For the generation view, each model answers the same 32 zero-shot AI/CS questions under a fixed system prompt; EDP is computed from the raw answers. Because generation uses an English system prompt with Chinese user queries, cross-model AvgEDP differences are descriptive and should not be interpreted as causal effects of instruction tuning. Model identifiers, prompts, collection settings, and known metadata gaps are listed in Appendix~\ref{app:setup}.

For the specialized-versus-general CSG comparison, we report 100,000-sample bootstrap confidence intervals and Welch tests, followed by Benjamini--Hochberg (BH) and Holm corrections over the nine models. We additionally fit term-clustered regressions on 1,077 reliably mapped strict pairs (990 specialized and 87 general; 850 term clusters), controlling for $\log(1+\text{Chinese-term frequency})$, mean Unicode-character length, and the character-length difference between pair members.

\definecolor{metahead}{RGB}{232,234,235}
\definecolor{pairhead}{RGB}{216,226,231}
\definecolor{pairbody}{RGB}{246,249,250}
\definecolor{diaghead}{RGB}{218,231,222}
\definecolor{diagbody}{RGB}{246,250,247}
\definecolor{groupband}{RGB}{241,242,242}
\begin{table*}[t]
\centering
\small
\setlength{\tabcolsep}{3.5pt}
\renewcommand{\arraystretch}{1.10}
\resizebox{\textwidth}{!}{
\begin{tabular}{@{}lcc>{\columncolor{pairbody}}r>{\columncolor{pairbody}}r>{\columncolor{pairbody}}r>{\columncolor{pairbody}}r>{\columncolor{diagbody}}r@{}}
\toprule
\multicolumn{3}{c}{\cellcolor{metahead}\textbf{Model information}} &
\multicolumn{4}{c}{\cellcolor{pairhead}\textbf{View A: Paired terminology choice}} &
\multicolumn{1}{c}{\cellcolor{diaghead}\textbf{View B: Open-ended conformity}} \\
\cmidrule(lr){1-3}
\cmidrule(lr){4-7}\cmidrule(l){8-8}
\cellcolor{metahead!55}\textbf{Model} &
\cellcolor{metahead!55}\textbf{Type} &
\cellcolor{metahead!55}\textbf{Size} &
\cellcolor{pairhead!60}\textbf{EPR (\%)} &
\cellcolor{pairhead!60}\textbf{CSG$_{\mathrm{gen}}$} &
\cellcolor{pairhead!60}\textbf{CSG$_{\mathrm{spec}}$} &
\cellcolor{pairhead!60}\textbf{Lift [95\% CI]} &
\cellcolor{diaghead!60}\textbf{AvgEDP} \\
\midrule
\rowcolor{groupband}\multicolumn{8}{@{}l}{\textit{Closed API systems}} \\
GPT-4o & API & -- & -- & -- & -- & -- & 4.360 \\
DeepSeek-V3 & API & -- & -- & -- & -- & -- & 4.195 \\
Claude-3.5-Sonnet & API & -- & -- & -- & -- & -- & 4.160 \\
\midrule
\rowcolor{groupband}\multicolumn{8}{@{}l}{\textit{Open-weight models (7--9B)}} \\
Meta-Llama-3-8B & Weights & 8B & 10.71 & -0.361 & -0.302 & 0.059 [-0.006, 0.132] & -- \\
Meta-Llama-3.1-8B-Instruct & Weights & 8B & -- & -- & -- & -- & 4.094 \\
Mistral-7B-v0.3 & Weights & 7B & 10.88 & -0.238 & -0.200 & 0.038 [-0.002, 0.080] & -- \\
Mistral-7B-Instruct-v0.3 & Weights & 7B & -- & -- & -- & -- & 3.137 \\
Yi-1.5-9B-Chat & Weights & 9B & 19.88 & -0.272 & -0.248 & 0.024 [-0.043, 0.092] & 3.103 \\
Qwen2.5-7B-Base & Weights & 7B & 13.20 & -0.387 & -0.404 & -0.018 [-0.090, 0.056] & 2.961 \\
Qwen2.5-7B-Instruct & Weights & 7B & 12.94 & -0.403 & -0.406 & -0.003 [-0.076, 0.071] & 2.714 \\
Gemma-2-9B-It & Weights & 9B & 5.57 & -0.603 & -0.555 & 0.049 [-0.040, 0.144] & 1.231 \\
\midrule
\rowcolor{groupband}\multicolumn{8}{@{}l}{\textit{Compact open-weight models ($<3$B)}} \\
Qwen2.5-0.5B-Instruct & Weights & 0.5B & 13.62 & -0.443 & -0.421 & 0.023 [-0.053, 0.098] & 4.166 \\
Qwen2.5-1.5B-Instruct & Weights & 1.5B & 11.83 & -0.442 & -0.423 & 0.019 [-0.054, 0.093] & 2.195 \\
TinyLlama-1.1B & Weights & 1.1B & 6.17 & -0.221 & -0.219 & 0.002 [-0.031, 0.035] & 1.771 \\
\bottomrule
\end{tabular}}
\caption{Main ChiEngMixBench results. View A uses the 1,167 strict single-replacement pairs and reports length-normalized likelihood preference between matched sentence realizations. View B reports mean clipped EDP on 32 generated answers. Lift is $\mathrm{CSG}_{\mathrm{spec}}-\mathrm{CSG}_{\mathrm{gen}}$ with a 95\% bootstrap CI. EDP measures conformity to the benchmark reference profile rather than general answer quality; no cross-view ranking is implied.}
\label{tab:main_bench}
\end{table*}

\subsection{Paired Terminology-Choice Preference}
\label{sec:paired_results}
Table~\ref{tab:main_bench} shows that the English-term realization is preferred on only 5.57--19.88\% of strict pairs. Thus, every evaluated open-weight model assigns higher length-normalized likelihood to the Chinese-counterpart realization on most matched items. This is a descriptive result about paired sentence realizations whose controlled contrast is terminology form, not a direct target-token probability or a claim that users would prefer the complete Chinese answer.

Seven of nine models have a positive raw specialized-minus-general CSG point estimate on the strict set, with lifts ranging from $-0.018$ to $0.059$. No strict-set bootstrap interval excludes zero, and no model is significant after BH or Holm correction at $q/p<0.05$. After controlling for term frequency and character length, the specialized coefficient remains significant for 0/9 models. On the broader 1,289-pair localized-edit sensitivity set, two raw intervals exclude zero but neither survives multiplicity correction. The warranted conclusion is therefore a weak directional tendency in some specifications, not a robust specialized-term effect.

\subsection{Open-Ended Generation Diagnostic}
AvgEDP varies substantially across models, but these values should be read as distances from the benchmark's mixed-style reference profile. For example, a low score may result from the English-ratio component even when an answer is fluent and factually adequate. Conversely, a high score does not establish that a user would prefer the complete response. We therefore do not rank models by AvgEDP as a general measure of naturalness or quality. Appendix~\ref{app:edp_audit} reports component outputs, clipping counts, and the formula-reproduction audit.

\subsection{Validation and Scope}
\label{sec:validation_scope}
Table~\ref{tab:validation_scope} in Appendix~\ref{app:validation_table} consolidates the analyses used to determine what the diagnostics do and do not support. These checks deliberately include negative and non-significant results.

\subsection{Human Evaluation}
\label{sec:human_eval}
The original calibration study contains 100 Likert-rated outputs. Three annotators independently completed separate sheets and could not see Machine\_Score, EDP, or one another's ratings. They were senior undergraduates in AI/CS-related programs at Tsinghua University, Beihang University, and HKUST (Guangzhou), with formal AI/CS training and Chinese--English technical-reading experience. Machine scores were merged only after annotation.

The high pre-consensus agreement in Table~\ref{tab:validation_scope} shows that the result is not an artifact of averaging the two closest ratings. Correlations are also stable under the all-rater mean, median, and closest-two mean (Appendix~\ref{app:human_details}). Nonetheless, EDP and the Likert rubric share constructs such as terminology form, mixing ratio, and verbosity. We therefore interpret their association as rubric-aligned calibration rather than independent proof of general response naturalness.

To test the broader interpretation directly, a separate blind study presented 77 pairs of complete responses and asked three annotators which answer was more idiomatic overall. Inter-annotator agreement was moderate, but EDP was near chance in predicting the majority preference. This result motivates the explicit scope boundary in Section~\ref{sec:framework}: EDP is a reference-profile diagnostic, not a substitute for holistic human evaluation.

\subsection{Additional Scope Checks}
\label{sec:scope_checks}
The Qwen2.5-7B Base/Instruct comparison does not support a general effect of instruction tuning on these metrics. The generation difference is non-significant, the CSG difference is essentially zero, and 28/32 Base outputs exhibit prompt-continuation artifacts. We therefore report this comparison only as a setup-sensitive Qwen observation and make no causal claim about RLHF.

For corpus contamination, the strict paired design controls much of the topic and local-context variation within each comparison. The remaining concern is differential training exposure to the original mixed form. As a bounded external audit, we collected 12 dated CSDN snippets covering six terms, with one pre-release and one post-release example per term, and evaluated Qwen2.5-7B-Instruct. The term-balanced pre/post difference is close to zero but highly uncertain. Because this pilot is not a temporal split of the candidate collection and the public model release date is not a verified training cutoff, it cannot establish the absence of contamination.

\section{Conclusion}
\label{sec:conclusion}

We introduced ChiEngMixBench, a resource and evaluation protocol for Chinese--English terminology mixing in AI/CS technical discourse. The release preserves a 1,706-item candidate collection with pair-quality labels, while the primary paired analysis uses 1,167 strict single-replacement contrasts. It separates preference between matched English-term and Chinese-counterpart sentence realizations from open-ended generation conformity, allowing these two behaviors to be inspected without treating either as a complete proxy for the other.

Across nine open-weight models, Chinese-counterpart realizations receive higher length-normalized likelihood on most strict pairs. Specialized terms show a positive raw CSG shift in seven models, but no strict-set interval excludes zero and the effect is not robust to multiple-comparison correction or frequency/length controls. For generated answers, EDP provides a transparent decomposition relative to the benchmark's reference profile. Reliability analyses support its use under the aligned rubric, while a strong ratio baseline and a separate A/B study delimit its interpretation: it is not a universal naturalness metric or a predictor of holistic human preference. We release the benchmark, quality audit, configuration, and analysis protocol to support reproducible study of terminology choice in professional multilingual settings.

\section*{Limitations}

\paragraph{Domain and population scope.}
ChiEngMixBench focuses on Chinese-matrix terminology mixing in AI/CS discussion. Practices in medicine, finance, or other language pairs may differ. The source platforms do not provide reliable author demographics, language histories, or expertise labels, so the corpus cannot establish how terminology preferences vary across user populations. We therefore evaluate conformity to the observed technical-discourse profile, not to a universal norm for Chinese speakers.

\paragraph{Web provenance and contamination.}
CSDN and Zhihu include reposted, search-optimized, and potentially model-generated content. Filtering and manual checks reduce obvious noise but cannot certify the provenance or expertise of every passage. Stable source-document and author-group identifiers were not retained for every released MCP, which prevents document-disjoint and author-disjoint splits. The paired design controls shared topic and prefix context, and our external temporal pilot finds no detectable pre/post drift, but neither check proves that benchmark text was absent from model pretraining.

\paragraph{Paired construction.}
The strict primary subset isolates one localized terminology contrast and therefore omits discourse-level decisions, multiple simultaneous switches, and interactions between terminology choice and answer content. The broader candidate collection includes cases where the target English abbreviation or term remains on the Chinese side, punctuation changes occur around the target span, or words outside the target terminology are rewritten; these items are labeled and excluded from the primary estimates. CSG compares length-normalized likelihoods of complete matched realizations; it is not a direct target-token probability and remains sensitive to model-specific tokenization. It should consequently be interpreted as within-pair terminology-choice evidence rather than a complete measure of code-mixing ability.

\paragraph{Platform style and accessibility.}
CSDN and Zhihu may amplify search-optimized writing, reposting, status-signaling, or platform-specific terminology habits, so conformity to this profile should not be treated as representative of all Chinese technical communities. English technical terms can improve efficiency for experienced practitioners while reducing accessibility for students, newcomers, and readers from adjacent domains. Optimizing models to mimic an insider style may therefore trade off against clarity for broader audiences; deployed systems should adapt terminology to users and context rather than maximize benchmark conformity indiscriminately.

\paragraph{Diagnostic calibration.}
EDP contains human-specified thresholds, weights, a reference centroid, and static terminology rules. These choices make it auditable and decomposable but task-specific. A ratio-only baseline is nearly as correlated with the original rubric-aligned Likert scores, and the 77-pair study shows that EDP does not predict holistic answer preference. Changes in terminology conventions will also require the reference profile and rules to be updated.

\paragraph{Model and generation settings.}
Closed APIs do not expose comparable log probabilities, so paired preference results are limited to open-weight models. Generation experiments use one fixed English system prompt and 32 questions; prompt language, decoding, endpoint revisions, and the prompt-continuation behavior of some Base models can confound Base/Instruct comparisons. Some older API logs do not retain exact endpoint revisions or timestamps. For these reasons, we do not draw a general conclusion about instruction tuning or RLHF.

\section*{Ethical Considerations}

\paragraph{Public technical text.}
The source material was collected from publicly accessible CSDN and Zhihu pages for non-commercial research. Public availability does not eliminate privacy or contextual-integrity concerns. The construction pipeline removed known direct identifiers from processed records and the benchmark does not require author names, profile links, avatars, or exact timestamps. Because automated scrubbing cannot guarantee removal of every indirect identifier, the revised release prioritizes de-identified benchmark items, derived annotations, and evaluation scripts rather than redistributing identifiable raw user pages.

\paragraph{Licensing and platform terms.}
The release documentation states the intended research license for the authors' processed annotations and code separately from rights that may remain with source platforms or original writers. Users are instructed not to reconstruct identities, redistribute raw platform content, or use the benchmark to profile individual authors. We also document the relevant platform Terms-of-Service boundary and remove source text when redistribution rights are unclear.

\paragraph{Human evaluation.}
The 100-item Likert evaluation and the 77-pair A/B study report only aggregate, anonymized judgments. Annotators completed the tasks independently and were blind to automated scores. We report their relevant academic and language background only at group level. The task asked for judgments about text, not sensitive personal information. No demographic or psychological inference is made from individual ratings.

\paragraph{Normative scope.}
ChiEngMixBench does not prescribe that Chinese communication should generally contain English terms, nor does it treat monolingual Chinese as inherently deficient. It evaluates conformity to terminology practices observed in a specific AI/CS setting. Such practices may exclude readers unfamiliar with English terminology; models deployed in practice should adapt to user preferences and accessibility needs rather than apply one community profile universally.

\paragraph{Research responsibility and AI assistance.}
The authors formulated the research questions, designed the benchmark and evaluation, curated and audited the data, ran the experiments, verified the numerical results and citations, interpreted the findings, and approved the final manuscript and release artifacts. AI assistants were used in a supporting role for language editing, drafting and restructuring prose, and code and debugging assistance during statistical and release-package checks. All AI-assisted suggestions and outputs were reviewed against the underlying data and source files. The authors retain responsibility for the study, claims, code, artifacts, and final text.

\bibliography{custom}

\appendix
\section{Validation and Scope Summary}
\label{app:validation_table}

Table~\ref{tab:validation_scope} summarizes the evidential boundary of the revised paper, including negative and non-significant checks retained for transparency.

\begin{table*}[t]
\centering
\footnotesize
\renewcommand{\arraystretch}{1.13}
\setlength{\tabcolsep}{4pt}
\begin{tabular}{p{0.20\textwidth}p{0.47\textwidth}p{0.27\textwidth}}
\toprule
Check & Result & Supported interpretation \\
\midrule
Original 100-item Likert reliability & Krippendorff's $\alpha$ (interval) $=0.9173$, 95\% CI $[0.8729,0.9481]$; ordinal $\alpha=0.7729$; ICC$(2,1)=0.9180$ and ICC$(2,k)=0.9711$. & Ratings are stable before consensus and do not depend on the closest-two rule. \\
Rubric-aligned score and ratio baseline & EDP vs. all-rater mean: Pearson $r=0.9667$, Spearman $\rho=0.7879$; ratio-only: $r=0.9642$, $\rho=0.8514$. EDP has lower MAE (0.3731 vs. 0.8101) and small incremental $\Delta R^2=0.0135$. & EDP is calibrated to the same mixed-style rubric; its advantage is decomposability and calibration, not a large correlation gain. \\
Blind 77-pair holistic A/B preference & Fleiss' $\kappa=0.554$; majority agreement with EDP is 38/74 non-ties (51.35\%, Wilson 95\% CI [40.18, 62.39]); AUC $=0.560$, $p=0.454$. & EDP does not reliably predict holistic human preference. \\
Specialized-term CSG & Strict set: 7/9 raw lifts are positive, but 0/9 bootstrap CIs exclude zero and 0/9 survive multiplicity correction or frequency/length controls. Localized-edit sensitivity: 2/9 raw CIs exclude zero and 0/9 survive correction. & Weak, specification-sensitive directional tendency only. \\
Pure-Chinese negative control & Relative to the mixed-style condition, EDP decreases for DeepSeek ($\Delta=-0.5277$, Holm $p=3.253\times10^{-5}$) and Gemini ($\Delta=-0.3397$, Holm $p=0.017$). & EDP responds to mixed-style conformity; it is not a general Chinese-answer quality metric. \\
Qwen Base/Instruct comparison & AvgEDP Instruct--Base $\Delta=-0.2475$, 95\% CI $[-0.9756,0.4575]$, $p=0.5085$; CSG $\Delta=0.000233$, $p=0.9096$. & No reliable evidence for a general instruction-tuning or RLHF effect. \\
External temporal pilot & Term-balanced post--pre CSG difference $=0.0098$, 95\% CI $[-0.1665,0.1815]$, $p=0.9385$ on 12 dated snippets. & No observed drift in this small audit; contamination is not ruled out. \\
\bottomrule
\end{tabular}
\caption{Validation and scope checks. ``Supported interpretation'' states the claim retained in the revised paper.}
\label{tab:validation_scope}
\end{table*}

\section{Dataset and Release Details}
\label{app:data}

\subsection{Benchmark Inventory}
The release preserves all 1,706 source-derived candidate pairs and 1,344 non-empty normalized target-term types. Each item provides a stable identifier, the two sentence variants, the recorded target, a descriptive term-group label, and a deterministic pair-quality label. The primary paired analysis contains 1,167 strict single-replacement pairs (1,079 specialized and 88 general) covering 934 normalized target terms. The 1,289-pair localized-edit sensitivity set contains the strict pairs plus 122 pairs with minor nearby surface edits.

The current public MCP release does not contain reliable upstream document or author-group identifiers. We therefore do not advertise document-disjoint or author-disjoint evaluation and do not reconstruct those identities by fuzzy matching. Future collection will assign irreversible document-group IDs before de-identification so that held-out grouping can be supported without releasing personal or source-platform identifiers.

\subsection{Construction Checks}
The post-collection audit uses exact target presence and character-level difference spans. Table~\ref{tab:pair_audit} reports all mutually exclusive labels. Two empty target fields (IDs 512 and 1184) are restored from the unambiguous English-side strings \emph{Hate} and \emph{Unseen} without changing sentence text. One duplicate strict item (ID 1420, duplicating the Licensing context in ID 648) is excluded from both analysis sets. The released full collection retains these records and documents each repair or exclusion.

\begin{table}[h]
\centering
\footnotesize
\setlength{\tabcolsep}{4pt}
\begin{tabular}{lr}
\toprule
Pair-quality label & Count \\
\midrule
Strict single replacement & 1,168 \\
Localized replacement with minor surface edits & 122 \\
Target retained on Chinese side & 294 \\
Non-local or multi-edit pair & 112 \\
Missing target metadata & 3 \\
Identical pair & 7 \\
\bottomrule
\end{tabular}
\caption{Deterministic pair-fidelity audit over the 1,706-item candidate collection. Counts precede duplicate removal; excluding one duplicate strict item yields the 1,167-pair primary set.}
\label{tab:pair_audit}
\end{table}

The strict rule requires the recorded target to occur on the English side, be absent on the Chinese side, and be covered by one changed block with at most two off-target changed characters. The localized rule allows up to three changed blocks and eight off-target changed characters. These string-level labels make the filter reproducible but do not replace linguistic validation; the sensitivity analysis tests whether admitting the localized cases changes the conclusion.

\subsection{External Temporal Pilot}
The temporal pilot is separate from the 1,706-item candidate collection. It contains 12 manually verified CSDN snippets covering six terms, with one example before and one example after the public release date of Qwen2.5-7B-Instruct (2024-09-19). The A/B variants differ only in the target term. Mean CSG is 0.099483 in the pre-release group and 0.109306 in the post-release group; EPR is 66.7\% in both. The term-balanced post--pre difference is 0.009822 (95\% CI $[-0.166539,0.181540]$, sign-flip $p=0.9385$). Excluding one page that discloses AIGC assistance gives $\Delta=0.0418$ (95\% CI $[-0.1688,0.2336]$, $p=0.8182$). This underpowered pilot is reported only as a bounded audit.

\section{EDP Configuration and Implementation Audit}
\label{app:edp_audit}

\subsection{Released Configuration}
Table~\ref{tab:edp_config} gives the exact component rules used in the reported EDP calculation. The implementation reads the historical aliases \texttt{ratio\_slope} and \texttt{switch\_slope} as the effective coefficients \texttt{ratio\_k} and \texttt{switch\_k}.

\begin{table}[h]
\centering
\footnotesize
\setlength{\tabcolsep}{3pt}
\renewcommand{\arraystretch}{1.14}
\begin{tabular}{@{}lp{0.72\columnwidth}@{}}
\toprule
\rowcolor{black!8}Component & Penalty rule \\
\midrule
Ratio & $P_i^{\mathrm{ratio}}=8\max\{0,\,0.01-r_i,\,r_i-0.283\}$ \\
\rowcolor{black!3}Switching & $P_i^{\mathrm{switch}}=8\max\{0,\,s_i-0.215\}$ \\
Semantic & $P_i^{\mathrm{semantic}}=d_i$ if $d_i\leq0.6557$; $3(d_i-0.6557)$ otherwise \\
\rowcolor{black!3}Terminology & $P_i^{\mathrm{term}}=\min\{1.6,\,0.4n_i\}$ \\
\bottomrule
\end{tabular}
\caption{Exact EDP component rules used for the reported scores.}
\label{tab:edp_config}
\end{table}

Here $r_i$ is the English-token ratio, $s_i$ is switching frequency, $d_i$ is cosine distance from the released reference centroid, and $n_i$ is the number of triggered terminology-form rules. The semantic distance uses \texttt{paraphrase-multilingual-MiniLM-L12-v2}; the fitted centroid vector is released separately. We expose the piecewise implementation rather than replacing it with a smoother post-hoc formula, because changing it would require rescoring and recalibration. The revised UTF-8 release validates the static terminology-rule file before scoring and records every triggered rule in the item-level output.

\subsection{Clipping and Formula Reproduction}
The historical results table reported the mean of item-level clipped scores. Table~\ref{tab:edp_audit} reproduces those values and shows why subtracting mean component penalties from 5 fails for high-penalty systems. Across all 12 models, the maximum absolute difference between the historical table mean and the audited implementation is 0.000975.

\begin{table*}[t]
\centering
\small
\setlength{\tabcolsep}{4pt}
\begin{tabular}{lrrr}
\toprule
Model & Reported AvgEDP & Zero-score items & Unclipped reconstruction \\
\midrule
Llama-3.1-8B-Instruct & 4.0944 & 0/32 & 4.0944 \\
Qwen2.5-0.5B-Instruct & 4.1659 & 1/32 & 3.9509 \\
Qwen2.5-1.5B-Instruct & 2.1950 & 15/32 & -1.2106 \\
Qwen2.5-7B-Base & 2.9613 & 5/32 & 2.6919 \\
Qwen2.5-7B-Instruct & 2.7138 & 12/32 & 0.0097 \\
TinyLlama-1.1B & 1.7713 & 19/32 & -2.7175 \\
DeepSeek-V3 & 4.1947 & 1/32 & 3.9806 \\
Claude-3.5-Sonnet & 4.1597 & 0/32 & 4.1603 \\
GPT-4o & 4.3603 & 0/32 & 4.3603 \\
Mistral-7B-Instruct-v0.3 & 3.1369 & 8/32 & 1.2728 \\
Gemma-2-9B-It & 1.2309 & 23/32 & -3.8653 \\
Yi-1.5-9B-Chat & 3.1031 & 3/32 & 2.8400 \\
\bottomrule
\end{tabular}
\caption{EDP implementation audit over 32 answers per model. ``Unclipped reconstruction'' is $5-\sum_k\overline{P_k}$ and is not the reported score.}
\label{tab:edp_audit}
\end{table*}

\section{Models, Prompts, and Reproducibility}
\label{app:setup}

\subsection{Evaluated Models}
The paired analysis includes Gemma-2-9B-It, Meta-Llama-3-8B, Mistral-7B-v0.3, Qwen2.5-0.5B-Instruct, Qwen2.5-1.5B-Instruct, Qwen2.5-7B-Base, Qwen2.5-7B-Instruct, TinyLlama-1.1B, and Yi-1.5-9B-Chat \citep{dubey2024llama,hui2024qwen2,zhang2024tinyllama}. The generation analysis uses Meta-Llama-3.1-8B-Instruct and Mistral-7B-Instruct-v0.3 rather than the corresponding paired-view checkpoints, and additionally includes GPT-4o, Claude-3.5-Sonnet, and DeepSeek-V3 \citep{hurst2024gpt,deepseek2024v3}. Available repository identifiers and raw-result filenames are provided in the release manifest; historical revisions that were not retained are explicitly marked as unavailable.

\subsection{Generation Prompt}
All generation systems received the same 32 AI/CS questions and the following system message:
\begin{quote}
\small
\textit{You are an AI assistant helping a computer science researcher. Please answer the following technical question concisely, maintaining a professional academic tone. Do not over-explain basic concepts.}
\end{quote}
The system message is in English while user questions are in Chinese. This is a fixed data-collection condition, not a theoretically privileged prompt. It can confound language behavior and is not used to support a causal claim about instruction tuning. The 32 questions cover computer vision, natural-language processing, and reinforcement learning.

Where generation metadata were retained, the API scripts used temperature 0.7 and \texttt{max\_tokens=2048}. Some older C1--C3 logs do not preserve endpoint revisions or collection timestamps; those comparisons are treated as descriptive, and the raw outputs are released. The C4 negative-control run records its settings separately. Open-weight probability probing was run from fixed model weights, and item-level scores are released so aggregate results can be recomputed.

\subsection{Recomputation}
The software package includes separate commands for (i) MCP likelihood scoring, (ii) EDP component scoring, and (iii) statistical aggregation. The package README specifies input schemas and exact command-line arguments. Each aggregate table is generated from item-level JSON/CSV rather than copied from the manuscript. Release checksums identify the data, configuration, scripts, and table artifacts used for the paper.

\section{Full CSG Statistical Results}
\label{app:csg_stats}

\begin{table*}[t]
\centering
\footnotesize
\setlength{\tabcolsep}{3.2pt}
\begin{tabular}{lrrrrrr}
\toprule
Model & Lift & 95\% bootstrap CI & Raw $p$ & BH $q$ & Holm $p$ & Raw CI excludes 0 \\
\midrule
Gemma-2-9B-It & 0.0486 & [-0.0404, 0.1439] & 0.3050 & 0.8177 & 1.0000 & No \\
Llama-3-8B & 0.0592 & [-0.0056, 0.1317] & 0.0946 & 0.4256 & 0.7566 & No \\
Mistral-7B-v0.3 & 0.0381 & [-0.0017, 0.0797] & 0.0714 & 0.4256 & 0.6429 & No \\
Qwen2.5-0.5B & 0.0226 & [-0.0529, 0.0980] & 0.5606 & 0.8177 & 1.0000 & No \\
Qwen2.5-1.5B & 0.0191 & [-0.0540, 0.0928] & 0.6131 & 0.8177 & 1.0000 & No \\
Qwen2.5-7B-Instruct & -0.0026 & [-0.0761, 0.0715] & 0.9462 & 0.9462 & 1.0000 & No \\
Qwen2.5-7B-Base & -0.0177 & [-0.0899, 0.0556] & 0.6360 & 0.8177 & 1.0000 & No \\
TinyLlama-1.1B & 0.0018 & [-0.0313, 0.0353] & 0.9177 & 0.9462 & 1.0000 & No \\
Yi-1.5-9B-Chat & 0.0239 & [-0.0428, 0.0917] & 0.4861 & 0.8177 & 1.0000 & No \\
\bottomrule
\end{tabular}
\caption{Specialized-minus-general CSG comparisons on the 1,167 strict pairs. BH and Holm corrections are applied over the nine model-wise Welch tests.}
\label{tab:csg_full}
\end{table*}

\subsection{Frequency and Length Controls}
The primary control subset contains 1,077 strict pairs with exact or normalized dictionary mappings (990 specialized and 87 general; 850 term clusters). The outcome is item-level CSG. Predictors are the specialized indicator, z-scored $\log(1+\text{Chinese-term frequency})$, mean Unicode-character length, and A-minus-B character-length difference. Confidence intervals use term-clustered standard errors, and model-wise $p$-values are Holm-adjusted.

\begin{table*}[t]
\centering
\footnotesize
\setlength{\tabcolsep}{4pt}
\begin{tabular}{lrrr}
\toprule
Model & Adjusted coefficient & 95\% CI & Holm $p$ \\
\midrule
Gemma-2-9B-It & 0.0286 & [-0.0645, 0.1217] & 1.0000 \\
Llama-3-8B & 0.0415 & [-0.0270, 0.1099] & 1.0000 \\
Mistral-7B-v0.3 & 0.0257 & [-0.0207, 0.0721] & 1.0000 \\
Qwen2.5-0.5B & 0.0019 & [-0.0818, 0.0856] & 1.0000 \\
Qwen2.5-1.5B & 0.0107 & [-0.0757, 0.0971] & 1.0000 \\
Qwen2.5-7B-Instruct & -0.0054 & [-0.0913, 0.0805] & 1.0000 \\
Qwen2.5-7B-Base & -0.0173 & [-0.1029, 0.0683] & 1.0000 \\
TinyLlama-1.1B & 0.0008 & [-0.0345, 0.0361] & 1.0000 \\
Yi-1.5-9B-Chat & 0.0174 & [-0.0576, 0.0923] & 1.0000 \\
\bottomrule
\end{tabular}
\caption{Controlled coefficient of the specialized indicator on the 1,077-pair strict subset.}
\label{tab:csg_controls}
\end{table*}

The 1,289-pair localized-edit sensitivity set yields raw bootstrap intervals excluding zero for Llama-3-8B (lift $=0.0711$, 95\% CI $[0.0111,0.1376]$) and Mistral-7B-v0.3 (lift $=0.0552$, 95\% CI $[0.0152,0.0965]$), but neither survives BH or Holm correction. A stricter frequency-mapping sensitivity subset contains 709 pairs (646 specialized and 63 general) and likewise yields no Holm-significant controlled coefficient. Exact model-tokenizer controls were not run for the revised strict subset; the released script records this unavailable branch rather than treating it as completed evidence.

\section{Human Evaluation and Baselines}
\label{app:human_details}

\subsection{Original 100-Item Likert Study}
The three annotators independently rated 100 outputs on a five-point rubric concerning fluency, terminology form, mixing pattern, and verbosity. Automated-score columns were absent from their sheets and were merged after completion. Before any consensus aggregation, Krippendorff's interval $\alpha$ is 0.9173 (95\% bootstrap CI $[0.8729,0.9481]$), ordinal $\alpha$ is 0.7729, ICC$(2,1)$ is 0.9180, and ICC$(2,k)$ is 0.9711. Pairwise quadratic-weighted kappas are 0.8400, 0.9542, and 0.8862; pairwise Pearson correlations are 0.9218, 0.9586, and 0.9086.

\begin{table*}[t]
\centering
\small
\begin{tabular}{llrrrr}
\toprule
Predictor & Human target & Pearson $r$ & Spearman $\rho$ & MAE & Incremental $\Delta R^2$ \\
\midrule
EDP & All-rater mean & 0.9667 & 0.7879 & 0.3731 & 0.0135 \\
Ratio-only & All-rater mean & 0.9642 & 0.8514 & 0.8101 & -- \\
EDP & All-rater median & 0.9637 & 0.7980 & 0.3856 & 0.0124 \\
Ratio-only & All-rater median & 0.9623 & 0.8689 & 0.8191 & -- \\
EDP & Closest-two mean & 0.9532 & 0.7932 & 0.4268 & 0.0159 \\
Ratio-only & Closest-two mean & 0.9483 & 0.8571 & 0.8581 & -- \\
\bottomrule
\end{tabular}
\caption{Correlations of EDP (Machine Score) and the English-ratio-only baseline with alternative human aggregations ($N=100$).}
\label{tab:human_corr_full}
\end{table*}

The incremental $\Delta R^2$ column measures EDP added after the ratio-only score. Partial correlations are 0.4385, 0.4090, and 0.3969 for the mean, median, and closest-two targets, respectively. These are positive but modest. Across the all-rater mean, EDP is closer to the human score on 65 items, ratio-only on 3, with 32 ties; this comparison concerns numerical calibration on the shared scale, not general human preference.

\subsection{Expanded 77-Pair A/B Study}
Three annotators independently completed all 77 blind response pairs. Fifty items are unanimous and 27 have a 2:1 split; mean itemwise agreement is 0.766 and Fleiss' $\kappa=0.554$. The majority labels are A for 48 items, B for 26, and tie for 3. On the 74 non-ties, EDP agrees with the majority on 38 (51.35\%, Wilson 95\% CI $[40.18,62.39]$). The corresponding AUC is 0.560 and the test against chance gives $p=0.454$. We therefore do not use this study as positive validation of holistic preference.

\section{Negative Controls and Comparison Boundaries}
\label{app:scope_details}

\subsection{Pure-Chinese Control}
For DeepSeek and Gemini, we added a C4 condition requesting a fully Chinese answer on the same 32 questions. Relative to the default mixed-style C1 condition, EDP decreases from 4.2201 to 3.6925 for DeepSeek ($\Delta=-0.5277$, 95\% CI $[-0.7154,-0.3462]$, Holm $p=3.253\times10^{-5}$) and from 4.1779 to 3.8382 for Gemini ($\Delta=-0.3397$, 95\% CI $[-0.5527,-0.1290]$, Holm $p=0.017$). This behavior is expected for a mixed-style diagnostic and also demonstrates why EDP must not be interpreted as a general quality metric for Chinese responses.

\subsection{Qwen Base/Instruct Comparison}
On the 32 generation prompts, Qwen2.5-7B Instruct minus Base has $\Delta=-0.2475$ in AvgEDP (95\% CI $[-0.9756,0.4575]$, paired $p=0.5085$). The paired CSG difference is 0.000233 (95\% CI $[-0.003784,0.004264]$, $p=0.9096$). Moreover, 28/32 Base generations display prompt-continuation artifacts. The comparison is therefore insufficient for a general conclusion about instruction tuning or RLHF.

\section{Artifact Checklist}
The revised release is organized around: (i) the complete 1,706-item collection with pair-quality labels and a schema; (ii) explicit ID lists and materialized files for the 1,167-pair primary and 1,289-pair sensitivity sets; (iii) the terminology mapping and UTF-8-validated rule list; (iv) model-specific item-level likelihood outputs aligned to both analysis sets; (v) the 32 generation prompts and raw outputs where redistribution is permitted; (vi) the EDP centroid and scalar configuration; (vii) human-evaluation protocols and anonymized ratings; (viii) analysis scripts for pair audit, confidence intervals, corrections, controls, and IAA; and (ix) a manifest with checksums. Source-platform identifiers, URLs, timestamps, and identifiable raw user pages are excluded from the public benchmark package.

\end{document}